\documentclass[10pt,twocolumn,letterpaper]{article}

\usepackage{wacv}              

\usepackage{graphicx}
\usepackage{amsmath}
\usepackage{amssymb}
\usepackage{booktabs}
\usepackage[accsupp]{axessibility}

\usepackage[pagebackref,breaklinks,colorlinks]{hyperref}

\usepackage[capitalize]{cleveref}
\crefname{section}{Sec.}{Secs.}
\Crefname{section}{Section}{Sections}
\Crefname{table}{Table}{Tables}
\crefname{table}{Tab.}{Tabs.}

\def\confName{WACV}
\def\confYear{2025}

\usepackage{orcidlink}

\usepackage{colortbl}
\usepackage{pifont}
\usepackage{multirow}
\usepackage{sansmath}
\newcommand{\method}{LoSA\xspace}
\newcommand{\cmark}{\ding{51}}%
\newcommand{\xmark}{\ding{55}}%
\newcommand\blfootnote[1]{%
  \begingroup
  \renewcommand\thefootnote{}\footnote{#1}%
  \addtocounter{footnote}{-1}%
  \endgroup
}

\begin{document}

\title{\method: \underline{Lo}ng-\underline{S}hort-range \underline{A}dapter for Scaling End-to-End Temporal Action Localization}

\author{Akshita Gupta*$^{1,2,3}$, Gaurav Mittal*$^1$, Ahmed Magooda$^1$, Ye Yu$^1$,  Graham W.~Taylor$^{2,3}$,
Mei Chen$^1$ \\
$^{1}$Microsoft, $^{2}$University of Guelph, $^{3}$ Vector Institute for AI
}

\maketitle

\begin{abstract}
  Temporal Action Localization (TAL) involves localizing and classifying action snippets in an untrimmed video. The emergence of large video foundation models has led RGB-only video backbones to outperform previous methods needing both RGB and optical flow modalities. Leveraging these large models is often limited to training only the TAL head due to the prohibitively large GPU memory required to adapt the video backbone for TAL. To overcome this limitation, we introduce \method, the first memory-and-parameter-efficient backbone adapter designed specifically for TAL to handle untrimmed videos. 
  \method specializes for TAL by introducing \textbf{Lo}ng-\textbf{S}hort-range \textbf{A}dapters that adapt the intermediate layers of the video backbone over different temporal ranges. These adapters run parallel to the video backbone to significantly reduce memory footprint. 
  \method also includes Long-Short-range Gated Fusion that strategically combines the output of these adapters from the video backbone layers to enhance the video features provided to the TAL head. Experiments show that \method significantly outperforms all existing methods on standard TAL benchmarks, THUMOS-14 and ActivityNet-v1.3, by scaling end-to-end backbone adaptation to billion-parameter-plus models like VideoMAEv2~(ViT-g) and leveraging them beyond head-only transfer learning. 
\end{abstract}

\vspace{-0.7cm}

\blfootnote{$*$ Equal Contribution. This work was done as Akshita’s internship project at Microsoft.}

\vspace{-2mm}
\section{Introduction}
\label{sec:intro}
\begin{figure*}[h]
    \centering
    \includegraphics[width=0.85\textwidth]{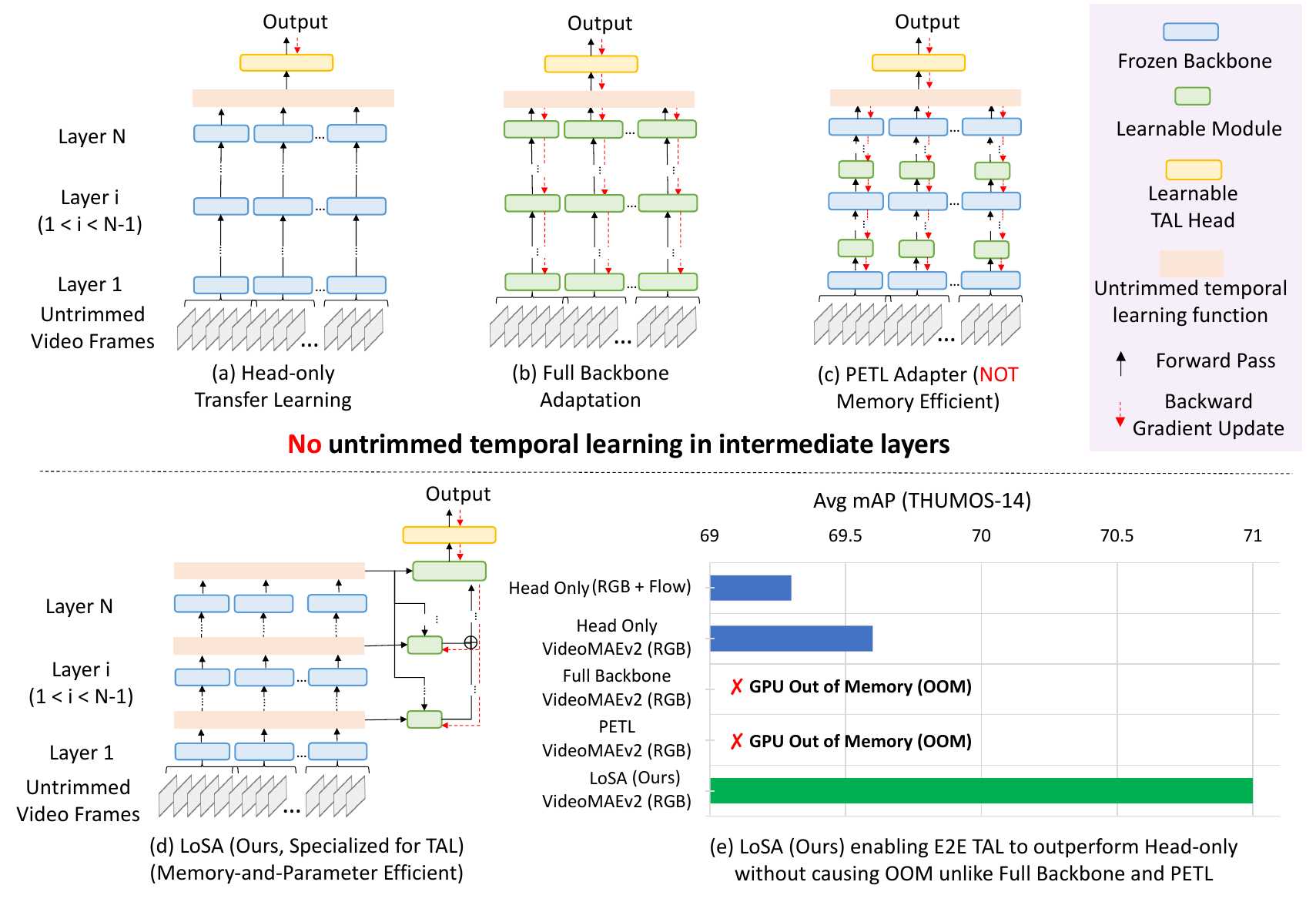}
    \vspace{-4mm}
    \caption{TAL Training Strategies/Performance.~\textbf{(a) Head-only Transfer Learning}: Untrimmed video frames processed as independent set of clips by the frozen backbone, features concatenated after last layer, and fed to learnable TAL head.~\textbf{(b) Full-backbone Transfer Learning}: Untrimmed video frames processed as independent set of clips by a learnable backbone, features concatenated after last layer, and fed to learnable TAL head.~\textbf{(c) Parameter-Efficient Transfer Learning~(PETL):} Untrimmed video frames processed as independent set of clips by a frozen backbone fitted with learnable adapter modules, features concatenated after last layer, and fed to learnable TAL head. Gradients backpropagate through entire backbone making PETL adapters parameter-efficient but not memory efficient. No untrimmed temporal learning in intermediate layers in (a-c).~\textbf{(d) \method~(Ours):} 
    Untrimmed video frames processed jointly at each intermediate layer, enabling untrimmed temporal learning by long- and short-range adapters~(green) to obtained TAL-enhanced features, and fed to learnable TAL head.
    No gradient backpropagating through backbone making \method both memory and parameter efficient. \textbf{(e)} On VideoMAEv2~(ViT-g) with THUMOS-14, only \method~(d) can perform end-to-end TAL while full backbone (b) and PETL (c) leads to GPU Out of Memory error, thereby significantly outperforming head-only (a).}
    \vspace{-6mm}
    \label{fig:concept}
\end{figure*}

Temporal Action Localization~(TAL) refers to localizing and classifying action snippets in an untrimmed~(arbitrarily long) video. TAL is crucial for applications in video indexing/search, surveillance, responsible AI, and robotics~\cite{wu2020not, mittal2024can}. 
Many TAL methods treat it as a downstream transfer learning task~\cite{zhang2022actionformer, rizve2023pivotal, wang2023videomae, xu2020g, yu2022batman}. Most works~\cite{zhang2022actionformer, lin2021learning, gupta2024open} perform head-only transfer learning where a frozen video backbone, generally pretrained on a large corpus of trimmed~($<$30s) videos like Kinetics-600~\cite{carreira2018short, patravali2021unsupervised}, is employed to extract features from untrimmed videos~(Fig~\ref{fig:concept}a). These features are then concatenated and fed to a trainable head designed to perform TAL. In this context, while certain studies~\cite{zhang2022actionformer, shi2023tridet, tang2023temporalmaxer} have shown improved results using both RGB and optical flow features, advances in video foundation models have enabled recent works~\cite{wang2023videomae, wang2022internvideo} to employ models like ViT-g with over 1\,B parameters~\cite{zhai2022scaling} to surpass previous methods with RGB features alone. 
This is because the effectiveness of data and model scaling is able to offset the need for expensive optical flow estimation~(Fig~\ref{fig:concept}e). 

Since TAL is performed on untrimmed videos, and these foundation models are typically trained on trimmed videos, there can be a distribution shift between backbone features and the downstream TAL task~\cite{xu2021boundary, wang2023protege}. This can result in confusion near action boundaries and fragmented action snippets.
This also suggests that adapting the backbone of the video foundation models beyond head-only transfer learning could help to further improve performance. 
Meanwhile, the massive size of foundation models, along with the long sequence length of untrimmed videos, make backbone adaptation prohibitively expensive \wrt GPU memory. Some TAL methods~\cite{cheng2022tallformer, liu2022empirical, zhao2023re2tal, cheng2022stochastic} propose memory optimizations to support full backbone adaptation~(Fig~\ref{fig:concept}b), but they cannot operate at the scale of foundation models, which are expected to increase in size over time. 

Recently, Parameter-efficient Transfer Learning/Fine-Tuning (PETL/ PEFT) approaches~\cite{houlsby2019parameter, hu2021lora, sung2022lst, yu2023visual} emerged, motivated by the computational constraints of adapting billion-parameter foundation models for downstream tasks. 
However, existing approaches are ill-equipped to learn context in untrimmed videos over different temporal ranges~(Fig~\ref{fig:concept}c) which is crucial to correctly localize actions of diverse type and duration~\cite{idrees2017thumos, caba2015activitynet}. These methods are thus sub-optimal in adapting the base foundation model for TAL.

To overcome this challenge, we introduce \method, the first memory- and-parameter-efficient backbone adapter that is tailored for TAL and untrimmed videos to harness large video foundation models more effectively beyond head-only transfer learning~(Fig~\ref{fig:concept}d). \method comprises a series of lightweight Long-range and Short-range Adapters that are attached to the intermediate layers of the video backbone. With video being processed by the large video foundation model as a sequence of trimmed video clips, these adapters learn to adapt the intermediate layers of the video backbone by capturing long-term and short-term dependencies among the video frames respectively. This allows an improved long-range temporal learning of the untrimmed video at each intermediate layer~(Fig~\ref{fig:concept}d) while also capturing the fine-grained short-term temporal changes in the video, allowing for more effective localization of actions. To allow each intermediate layer to contribute directly
towards improving TAL, the adapters leverage cross-attention between intermediate layers and the last layer of the video backbone to transform the output of intermediate layers. 

The Long-range and Short-range Adapters run parallel to the video backbone and their outputs directly aggregate with the last layer features. This circumvents gradient backpropagation through the video backbone, which significantly reduces the memory footprint of adapting the backbone for TAL. To facilitate this aggregation, \method introduces Long-Short-range Gated Fusion with a learnable gating function to weigh the contribution of each intermediate layer and fuse them together with the output of the last layer to generate improved features for the TAL head. These TAL-enhanced features enable more accurate action boundaries compared to head-only transfer learning as evident from the superior performance in Fig~\ref{fig:concept}e.

We demonstrate \method's effectiveness in adapting video backbones for TAL on both transformer-based and CNN-based models including VideoMAEv2 (ViT-g) which has $>$1\,B parameters. Experiments on standard TAL datasets, THUMOS-14 and ActivityNet-v1.3, show that \method significantly outperforms all existing methods and PETL techniques by enabling end-to-end backbone adaptation of large video foundation models beyond head-only transfer learning.
In summary:
\vspace{-2mm}
\begin{enumerate}
    \setlength\itemsep{0em}
    \item We address the significant challenge in the TAL field of scaling end-to-end training by introducing \method, an innovative solution for TAL that is specifically tailored for untrimmed videos. %
    \item \method comprises a novel adapter design to enable memory-and-parameter efficiency for untrimmed videos by employing a series of lightweight Long-range and Short-range adapters that run parallel to the video backbone and a Long-Short-range Gated Fusion module to adaptively fuse the outputs from the 
    adapters to improve TAL.
    \item \method is capable of end-to-end backbone adaptation of $>$1\,B parameter video models beyond head-only transfer learning, establishing new SOTA for TAL. 
\end{enumerate}
\section{Related Work}
\label{sec:related_work}
\noindent {\bf Temporal Action Localization~(TAL).}
Most TAL approaches leverage RGB and optical flow features pre-extracted from a video backbone. Among these are two-stage methods~\cite{xu2020g, bai2020boundary, lin2018bsn, lin2019bmn, gong2020scale, zhao2020bottom}, which generate pre-defined action proposals and then classify them into action classes while regressing the actual action boundaries, and one-stage methods~\cite{lin2021learning, liu2022end, zhang2022actionformer, shi2023tridet, tang2023temporalmaxer}, which perform TAL in a single pass without separately generating action proposals. These approaches perform head-only transfer learning, treating the video backbone as frozen.~In spite of shallow training and leveraging relatively small backbones like I3D~\cite{carreira2017quo}, TSN~\cite{wang2016temporal}, and TSP~\cite{alwassel2021tsp}, they achieve competitive performance by using optical flow that enhances temporal understanding. However, optical flow estimation is computationally expensive, making it challenging to scale on increasingly large video datasets. Recently, large video foundation models~\cite{wang2023videomae, wang2022internvideo} have demonstrated superior performance on TAL using RGB features only. These are also limited to head-only transfer learning due to the prohibitively large GPU memory footprint for end-to-end training.

\noindent {\bf Backbone Adaptation approaches for TAL.}
There exist approaches~\cite{cheng2022tallformer, liu2022empirical, zhao2023re2tal, cheng2022stochastic} that attempt to adapt an RGB-only video backbone beyond head-only transfer learning to mitigate the need for optical flow. They do so via memory optimizations such as reducing spatial resolution~\cite{liu2022empirical}, channel activations~\cite{cheng2022stochastic}, feature caching~\cite{cheng2022tallformer}, and rewiring the backbone~\cite{zhao2023re2tal}. While they can operate on relatively small backbones like SlowFast-101~\cite{feichtenhofer2019slowfast}, ViT-B~\cite{dosovitskiy2020image}, and ResNet-50~\cite{he2016deep}, they fail to scale to the size of current visual foundation models with billions of parameters~\cite{wang2023videomae}. \method, with its memory- and parameter-efficient backbone adapter, mitigates this issue and enables backbone adaptation of RGB-only large video backbones beyond head-only transfer learning to outperform all existing methods.

\noindent {\bf Parameter-efficient Transfer Learning~(PETL).}
With the advent of large-language models~(LLMs) \cite{brown2020language, touvron2023llama}, parameter-efficient transfer learning/finetuning (PETL/PEFT)~\cite{hu2021lora, houlsby2019parameter} has emerged to reduce computational costs of finetuning LLMs on downstream tasks. Inspired by LLM-based PETL, vision-based PETL was developed to enable efficient transfer learning on visual tasks. Yet, most approaches are parameter-efficient but not memory-efficient~\cite{sung2022lst, yang2022aim} as their design causes gradient backpropagation through the backbone. Some more recent
methods~\cite{yin2023parameter, sung2022lst} 
attempt to address memory efficiency, but no existing method, to the best of our knowledge, is suited to handle untrimmed videos. \method is the first memory- and parameter-efficient approach that is designed for TAL. 

\section{Method}
\begin{figure*}[t]
    \centering
    \includegraphics[width=\textwidth]{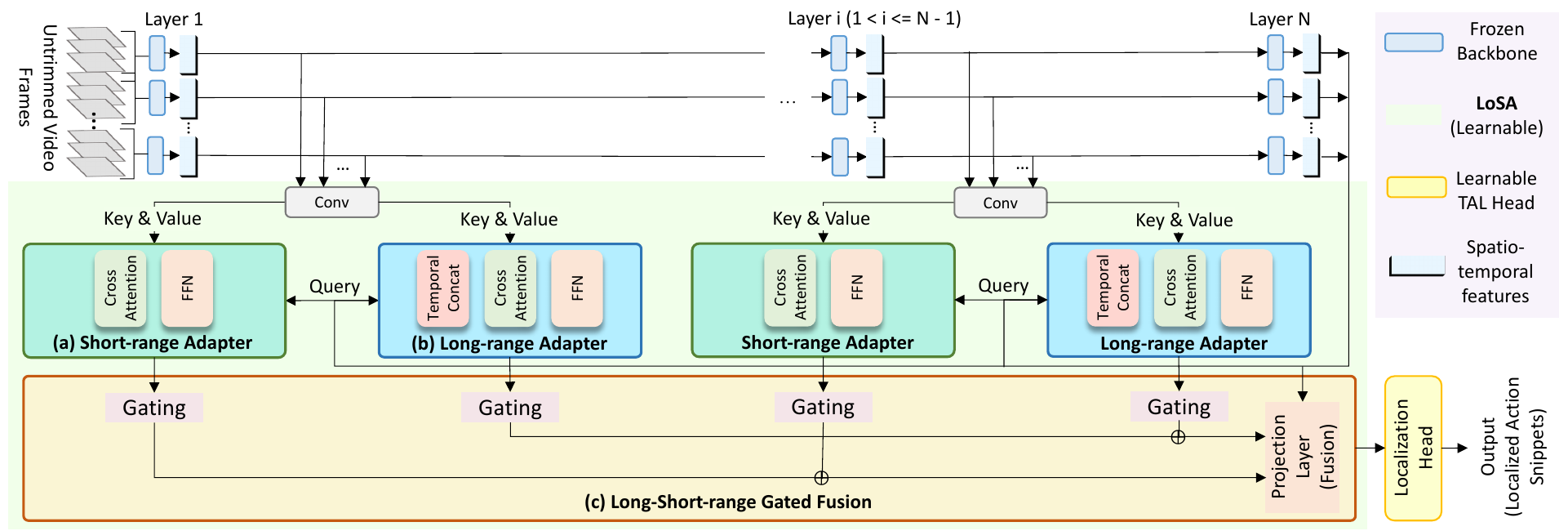}
    \vspace{-7mm}
    \caption{\textbf{\method Overview}: \method comprises a series of Long-range and Short-range Adapters that attach to the intermediate layers $1, \hdots, N-1$ of a video backbone. \textbf{(a)}~Each \textbf{Short-range Adapter} consists of a cross-attention module that uses the video clip-level spatio-temporal features of an intermediate layer as Query and the last layer temporally concatenated features as Key and Value. \textbf{(b)}~Similarly, each \textbf{Long-range Adapter} uses a cross-attention module to cross-attend the temporally concatenated long-range untrimmed video features of an intermediate layer as Query~(Q) and the last layer temporally concatenated features as Key~(K) and Value~(V). \textbf{(c)}~Finally, the \textbf{Long-Short-range Gated Fusion} module learns scaling parameters to gate the contribution of the Long-range and Short-range Adapters and combines them with the temporally concatenated last layer features via a projection layer to generate the TAL-enhanced features going into the learnable TAL head for outputting the localized action snippets.}
    \vspace{-4mm}
    \label{fig:model}
    \vspace{-3mm}
\end{figure*}
We describe the components of \method in the following subsections. Fig~\ref{fig:model} provides an overview of the model.

\subsection{Preliminaries}
Let $g$ denote a video backbone comprising $N$ layers, $f_1, \hdots, f_N$, defined as
$g = f_N (f_{N-1} (\hdots f_1(\mathbf{X}) \hdots))$, where $\mathbf{X}$ is the input to the backbone. 
Let $F_i$ be the feature representation obtained as the output from layer $f_i$. 
To adapt $g$ for TAL, $\mathbf{X}$ is an untrimmed video comprising an arbitrary number of frames. Since the existing video backbones are trained using trimmed video clips, we divide $\mathbf{X}$ into a sequence of $T$ clips ${x_1, \hdots, x_T}$ where each clip is a sequence of $T'$ frames such that $T'\ll$ total number of frames in the untrimmed video. Depending on the stride, the clips can be either overlapping or disjoint. We feed each clip $x_t~\forall~t \in \{1 \hdots T\}$ to the video backbone $g$ to generate a set of feature maps $F_i^{x_t}~\forall~i \in \{1 \hdots N\}, t \in \{1 \hdots T\}$ corresponding to all the layers of $g$. We assume each feature map $F_i^{x_t} \in \mathbb{R}^{T_i\times H_i \times W_i \times C_i}$ where $T_i$, $H_i$, $W_i$, and $C_i$ denote the temporal, height, width, and channel dimensions of the video clip features at intemediate layer $f_i$. Further, at each intermediate layer $f_i$, let $\mathbf{F^X_i} = \{F_i^{x_1}, \hdots, F_i^{x_T}\} \in \mathbb{R}^{TT_i \times {H_i \times W_i \times C_i}}$ be the concatenation of the feature maps along the temporal dimension. 

\subsection{Short-range Temporal Adapter}
Let $F_i^{x_t} \in \mathbb{R}^{T_i\times H_i \times W_i \times C_i}$ be an intermediate spatio-temporal feature map  from layer $f_i$ at temporal location $t$ where $i \in \{1 \hdots N-1\}$ and $t \in \{1 \hdots T\}$ obtained from feeding a trimmed video clip of the untrimmed video to the video backbone.
Since $T'\ll$ total number of frames in the untrimmed video, $F_i^{x_t}$ captures a short-range temporal context and provides a fine-grained temporal understanding in the local temporal neighborhood of the untrimmed video around its timestamp $t$. The video backbone processes each $F_i^{x_t}~\forall~t \in \{1 \hdots T\}$ independent of each other, which prevents the context of the full untrimmed video to influence the representation learning. So to perform transfer learning over these features for effective end-to-end TAL, \method comprises a series of Short-range Temporal Adapters~(Fig~\ref{fig:model}a) to adapt $F_i^{x_t}~\forall~t \in \{1 \hdots T\}, i \in \{1 \hdots N-1\}$ conditioned on all the short-range spatio-temporal features from the last layer, $F_N^{x_t}~\forall~t \in \{1 \hdots T\}$.

For this, we introduce a multi-headed cross-attention module in the Short-range Temporal Adapter at each intermediate layer $f_i~\forall~i \in \{1 \hdots N-1\}$ of the video backbone. We first employ a convolutional block at each layer $i \in \{1 \hdots N\}$ of the video backbone to process the spatial dimensions by transforming 
$F_i^{x_t} \in \mathbb{R}^{T_i\times H_i \times W_i \times C_i} 
\rightarrow {F'}_i^{x_t} \in \mathbb{R}^{T_i\times C_i}$.
For the cross-attention, we use ${F'}_i^{x_t}$ as Query and the feature set $\{{F'}_N^{x_1}, \hdots, {F'}_N^{x_T}\}$ from last layer $N$ as Key and Value. 
The cross-attention module outputs $FS^{x_t}_i \in \mathbb{R}^{T_i\times C_i}~\forall~t \in \{1 \hdots T\}, i \in \{1 \hdots N-1\}$, denoting the short-range features adapted for end-to-end TAL.

\subsection{Long-range Temporal Adapter}
While the Short-range Temporal Adapters enable features from trimmed video clips to be adapted at each temporal location for TAL for all intermediate layers, it is not sufficient for effectively localizing actions. Understanding the long-range temporal relationship among frames over the full untrimmed video is required to correctly identify the action boundaries and effectively distinguish between foreground and background. No existing method, including both full backbone adaptation and PETL~(Fig~\ref{fig:concept}b,c), incorporate a mechanism to capture this long-range temporal understanding in an untrimmed video directly at intermediate layers. To address this, \method comprises a series of Long-range Temporal Adapters~(Fig~\ref{fig:model}b) that learn to adapt the full temporally concatenated feature map sequence $\mathbf{F^X_i}$ jointly for end-to-end TAL at each intermediate layer $f_i~\forall~i \in \{1 \hdots N-1\}$. 

The Long-range Temporal Adapter adapts the feature sequence at a given intermediate layer conditioned on the feature sequence from last layer of the video backbone. For this, each Long-range Temporal Adapter consists of a cross-attention at each intermediate layer $f_i~\forall~i \in \{1 \hdots N-1\}$ of the video backbone. We feed the output of the convolutional block, concatenated along temporal dimension, $\mathbf{{F'}^X_i}$ as Query and $\mathbf{{F'}^X_N}$ as Key and Value to the cross-attention module. By doing so, we consider the last layer features as reference to cross-attend the intermediate long-range temporal features to adapt the latter into what directly improves the features used for TAL. The cross-attention module outputs $\mathbf{FL^X_i} \in \mathbb{R}^{TT_i\times C_i}~\forall~i \in \{1 \hdots N-1\}$,
representing the long-range features adapted for end-to-end TAL.

\subsection{Long-Short-range Gated Fusion}
\label{sec:loshrf}
The output of the adapters comprises feature representations at different temporal ranges and different intermediate layers of the video backbone. 
We maintain that these features from intermediate layers should function to enhance the temporal understanding of the last layer features during end-to-end training.
We therefore need to strategically fuse these representations to generate TAL-enhanced feature representations for the TAL head that are the most optimal in localizing actions. 
For this, after obtaining the short-range and long-range features from each intermediate layer, we introduce a Long-Short-range Gated Fusion module~(Fig~\ref{fig:model}c) to learn to strategically fuse these features along with the last layer features to obtain the TAL-enhanced features that are fed to the TAL head.

As shown in Fig~\ref{fig:model}c, the Long-Short-range Gated Fusion module comprises a short-range gating layer, $\mathsf{Gate}^{\mathsf{sh}}_i$, at each intermediate layer $f_1, \hdots, f_{N-1}$. $\mathsf{Gate}^{\mathsf{sh}}_i$ multiplies the temporally concatenated short-range features, obtained from the Short-range Adapter, at layer $f_i$, $\mathbf{FS^X_i} = \{{FS}_i^{x_1}, \hdots, {FS}_i^{x_T}\} \in \mathbb{R}^{TT_i\times C_i}$ with a learnable scaling parameter $p^{\mathsf{sh}}_i$ to compute the short-range contribution at intermediate layer $f_i$ as,
$
\mathsf{Gate}^{\mathsf{sh}}_i(\mathbf{FS^X_i}) = p^{\mathsf{sh}}_i\mathbf{FS^X_i}
$.
Next, we sum the short-range contributions over all intermediate layers as,
$
    \mathbf{FS^X} = \sum_{i=1}^{N-1} \mathsf{Gate}^{\mathsf{sh}}_i(\mathbf{FS^X_i}).\
$ We note that depending on the network architecture $T_i$ may vary in size across intermediate layers. To accommodate that and faciliate fusion, we use linear projection layers, as needed, to transform each $T_i \rightarrow T_N$, temporal dimension of the last layer $f_N$. 

Similarly, the Long-Short-range Gated Fusion module comprises a long-range gating layer, $\mathsf{Gate}^{\mathsf{lo}}_i$, at each intermediate layer $f_1, \hdots, f_{N-1}$. $\mathsf{Gate}^{\mathsf{lo}}_i$ multiplies the long-range features, obtained from the Long-range Adapter, at layer $f_i$, $\mathbf{FL^X_i} \in \mathbb{R}^{TT_i\times C_i}~\forall~i \in \{1 \hdots N-1\}$ with a learnable scaling parameter $p^{\mathsf{lo}}_i$ to compute the long-range contribution at intermediate layer $f_i$ as,
$
\mathsf{Gate}^{\mathsf{lo}}_i(\mathbf{FL^X_i}) = p^{\mathsf{lo}}_i\mathbf{FL^X_i}
$.
Next, we sum the long-range contributions over all intermediate layers as,
$
    \mathbf{FL^X} = \sum_{i=1}^{N-1} \mathsf{Gate}^{\mathsf{lo}}_i(\mathbf{FL^X_i}).\
$
Similar to short-range gating, we use linear projection layers, as needed, to make temporal dimensions at different layers consistent with that of last layer $f_N$, \ie, $T_N$.

After obtaining $\mathbf{FS^X}$ and $\mathbf{FL^X}$, we combine them with the last layer features $\mathbf{F_N^X}$ by addition and finally feed the concatenation of the resulting set of features to a linear projection layer $\mathsf{Proj}: \mathbb{R}^{T_N \times 2C} \rightarrow \mathbb{R}^{T_N \times C}$ to obtain the TAL-enhanced features, $\mathbf{FT^{X}_N}$, to be fed to the TAL head as, 
\begin{equation}
\small
    \mathbf{{FS'}^X} = \mathbf{FS^X} + \mathbf{F_N^X},\quad \mathbf{{FL'}^X} = \mathbf{FL^X} + \mathbf{F_N^X},
\end{equation}
\vspace{-4mm}
\begin{equation}
    \mathbf{FT^{X}_N} = \mathsf{Proj}([\mathbf{{FS'}^X}, \mathbf{{FL'}^X}]).\
\end{equation}

By doing so, we consider the short-range and long-range contribution of intermediate layers as a residual contribution to the original last layer features. To mathematically enforce that, we perform zero initialization on the scaling parameters $\{p^{\mathsf{sp}}_i\}^{N-1}_{i=0}$ and $\{p^{\mathsf{tem}}_i\}^{N-1}_{i=0}$ of gating layers $\{\mathsf{Gate}^{\mathsf{sp}}_i\}^{N-1}_{i=0}$ and $\{\mathsf{Gate}^{\mathsf{tem}}_i\}^{N-1}_{i=0}$ respectively. This ensures that when training starts, the TAL-enhanced features entering the TAL head are effectively the last layer's original features, providing a stable baseline for leveraging video backbone $g$. We finally feed $\mathbf{FT^{X}_N}$ to TAL head for generating the localized action snippets as output~(Fig~\ref{fig:model}, bottom-right).

\subsection{Enabling Memory-and-Parameter Efficiency}
The unique adapter design of \method enables temporal video understanding over the full untrimmed video at each intermediate layer of the video backbone during end-to-end training~(Fig~\ref{fig:concept}d). This allows the video backbone, originally trained on trimmed videos, to improve its understanding of the untrimmed video over long-range and short-range temporal context. This is unlike any existing end-to-end TAL method, including both full backbone adaptation and PETL~(Fig~\ref{fig:concept}b,c), where there is no mechanism to capture this long-range and short-range temporal understanding of an untrimmed video directly at intermediate layers. A crucial challenge in enabling untrimmed temporal learning at each intermediate layer during end-to-end training is the prohibitively large GPU memory footprint. This is because for TAL, each untrimmed video in a training batch involves processing several video clips together. The memory issue is further aggravated when leveraging $>$1\,B parameter video backbones. \method's adapter design, performing cross-attention with last layer features, allows the Long-range and Short-range Temporal Adapters to run parallel to the video backbone~(Fig~\ref{fig:model}). This makes the design memory-and-parameter efficient and circumvents backpropagating gradients through the video backbone, thereby significantly reducing the GPU memory footprint.

\section{Experiments}
\noindent{\bf Datasets.}~We evaluate \method on the two standard datasets for TAL: THUMOS-14~\cite{idrees2017thumos} and ActivityNet-v1.3~\cite{caba2015activitynet}. THUMOS-14 has 20 action classes. Following existing methods~\cite{zhang2022actionformer, lin2021learning, wang2023videomae}, we use the 200 untrimmed videos in the validation set for training and test on a set of 212 test videos. ActivityNet-v1.3 has 200 action classes. We use the 10,024 videos from the training set for training and use the 4,926 videos from the validation set for testing.

\noindent{\bf Model Backbones.}~We consider three video backbones: SlowFast-101~\cite{feichtenhofer2019slowfast},~VideoMAEv2~(ViT-Base)~\cite{wang2023videomae},~and VideoMAEv2~(ViT-g)~\cite{wang2023videomae}. VideoMAEv2~(ViT-g) has $\sim$1.01 billion parameters and is among the recent large video foundation models. It achieves SOTA on several video benchmarks including TAL. We select VideoMAEv2 (ViT-g) to demonstrate the scaling limitations of existing methods as well as \method's effectiveness in overcoming those limitations. We select SlowFast-101 and VideoMAEv2 (ViT-Base) as these models are widely used in the TAL literature. This further helps to evaluate the effectiveness of \method on different families of model architecture with SlowFast-101 being CNN-based~\cite{krizhevsky2012imagenet} and VideoMAEv2~(ViT-Base) being a Transformer~\cite{vaswani2017attention}-based model. 

\noindent{\bf Implementation Details.}~We train \method by feeding RGB-frames (similar to existing end-to-end TAL methods) as input to the different video backbones with an initial LR of 1e-4 for THUMOS-14 and 1e-3 for ActivityNet using a cosine annealing, warmup of 5 epochs, and AdamW~\cite{loshchilov2018decoupled} optimizer. We use Actionformer as the TAL head with max sequence length of 576 frames at $224 \times 224$ spatial resolution with $T'=16$ frames. We use temporally-consistent spatial augmentation involving random resizing and cropping and autoaugment~\cite{cubuk2019autoaugment}. We attach Short-range and Long-range Adapters to all the intermediate layers with $n_{\text{heads}} = 4$. 
Please refer to the supplementary for additional details. 

\subsection{Head-only vs.~End-to-End TAL Training}
\begin{table}[h]
    \vspace{-5mm}
    \centering
    \resizebox{\columnwidth}{!}{%
    \begin{tabular}{l|c|c|c|c}
    \hline
    
    \hline
       \multirow{2}{*}{Method}  & \multirow{2}{*}{Backbone} & \multirow{2}{*}{\shortstack[c]{End-to-End\\Adaptation}} & 
       \multirow{2}{*}{\shortstack[c]{GPU}}  & \multirow{2}{*}{\shortstack[c]{Avg mAP ($\uparrow$)}}  \\ 
       & &  & & \\
        \hline 
        
        \hline
        Head-only &\multirow{3}{*}{\shortstack[c]{SlowFast-101}}  & \xmark & 1.8 GB & 55.1 \\
        Full Backbone & & \cmark & 14 GB & 56.4 \\ 
        \method & & \cmark & 3.5 GB & \textbf{58.2} \\ \hline 
        
        \hline
        Head-only &\multirow{3}{*}{\shortstack[c]{VideoMAEv2\\(ViT-g)}}  & \xmark  & 2.4 GB & 69.6 \\
        Full Backbone & & \cmark & OOM & - \\ 
        \method & & \cmark  & 40.6 GB & \textbf{71.0} \\ \hline

        \hline
    \end{tabular}
    }
    \vspace{-2mm}
    \caption{Comparison of \method with other TAL training strategies for TAL on THUMOS-14. E2E Adaptation is \cmark~when backbone adaptation~(end-to-end training) happens along with learning the TAL head. GPU - peak GPU memory occupied overall by training for batch size of 1 on an A100 GPU. OOM - out of GPU memory error when even batch size of 1 cannot fit in GPU. Avg mAP is `-' when there is OOM as training fails to run. On 1.01B parameter VideoMAEv2~(ViT-g), only \method, by being both memory and parameter efficient, is able to perform end-to-end backbone adaptation beyond Head-only and achieve superior Avg mAP. On SlowFast-101, \method can outperform all TAL training strategies.}
    \vspace{-3mm}
    \label{tab:intro_petl}
\end{table}
Table~\ref{tab:intro_petl} provides a comparison of \method with different training strategies available for TAL~(as shown in Fig~\ref{fig:concept}). We experiment using THUMOS-14 on SlowFast-101 and VideoMAEv2~(ViT-g) to show the comparison on backbones with sizes at different orders of magnitude. We can observe that for both backbones, \method significantly outperforms head-only transfer learning~(Fig~\ref{fig:concept}a) by $3.1\%$ and $1.4\%$ respectively on Avg mAP. \method also outperforms full backbone adaptation on SlowFast-101 by $1.8\%$ Avg mAP. Full backbone adaptation~(Fig~\ref{fig:concept}b) results in GPU Out of Memory error (OOM) on VideoMAEv2~(ViT-g) due to the backbone having 1.02 billion parameters, which prevents even a batch of one training sample to fit in an A100 GPU. 

\subsection{Different Adapter designs for TAL}
\begin{table}[h]
    \vspace{-4mm}
    \centering
    \resizebox{\columnwidth}{!}{%
    \begin{tabular}{l|c|c|c|c|c|c}
    \hline
    
    \hline
       \multirow{2}{*}{Method}  & \multirow{2}{*}{Backbone} & \multirow{2}{*}{\shortstack[c]{End-to-End\\Adaptation}} & \multicolumn{2}{c|}{Backbone Parameters} & 
       \multirow{2}{*}{\shortstack[c]{GPU}}  & \multirow{2}{*}{\shortstack[c]{\shortstack[c]{Avg\\mAP ($\uparrow$)}}}  \\ \cline{4-5}
       &  & & Full & Learnable & & \\
        \hline 
        
        \hline
        Full Backbone &\multirow{5}{*}{\shortstack[c]{SlowFast-101}}  & \cmark & \multirow{5}{*}{\shortstack[c]{62M}}&  62M & 14 GB & 56.4 \\
        ST-Adapter*~\cite{pan2022st} & & \cmark & &  8M (-87\%) & 10 GB & 53.2 
        \\
        AIM*~\cite{yang2022aim} & & \cmark & &  10M (-83\%) & 12 GB & 54.0 
        \\
        \method & & \cmark & &  12M~(-80\%) & 3.5 GB & \textbf{58.2} \\ \hline 
        
        \hline
        Full Backbone &\multirow{5}{*}{\shortstack[c]{VideoMAEv2\\(ViT-g)}}  & \cmark &  \multirow{5}{*}{\shortstack[c]{1012M}} & 1012M & OOM & - \\
        ST-Adapter*~\cite{pan2022st} & & \cmark & &  88M (-91\%) & OOM & - \\
        AIM*~\cite{yang2022aim} & & \cmark & &  92M (-90\%) & OOM & - \\
        \method & & \cmark & &  143M (-86\%) & 40.6 GB & \textbf{71.0} \\ \hline

        \hline
    \end{tabular}
    }
    \vspace{-2mm}
    \caption{Comparison of \method with other training strategies for TAL on THUMOS-14. E2E Adaptation is \cmark~when backbone adaptation~(end-to-end training) happens along with learning the TAL head. GPU - peak GPU memory occupied overall by training for batch size of 1 on an A100 GPU. OOM - out of GPU memory error when even batch size of 1 cannot fit in GPU. Avg mAP is `-' when there is OOM as training fails to run. Learnable column includes percentage reduction in parameters \wrt Full column in parentheses. On 1.01 billion parameter VideoMAEv2~(ViT-g), only \method, by being both memory and parameter efficient, is able to perform end-to-end backbone adaptation and achieve superior Avg mAP. On SlowFast-101, where all training strategies can operate, \method can still outperform all the baselines. *Repurposed for TAL.}
    \vspace{-3mm}
    \label{tab:petl_comparison}
\end{table}

Since there exists no previous adapter-based PETL method for TAL, we re-purpose some of the existing PETL approaches to work for TAL.
We consider ST-Adapter and AIM for comparison. For SlowFast-101, \method significantly outperforms all existing adapter-based methods by at least $1.4\%$ Avg mAP, highlighting the importance of the design of \method's Long-Short-range Adapter tailored specifically for TAL. On VideoMAEv2~(ViT-g), the original implementation of ST-Adapter and AIM leads to an OOM error because, as Table~\ref{tab:petl_comparison} and Fig~\ref{fig:concept}c show, while their learnable parameter count is significantly less than the full backbone, they are not memory efficient, making them unscalable to billion parameter models like VideoMAEv2~(ViT-g). \method works on VideoMAEv2~(ViT-g) because it is both memory and parameter efficient. 
Additionally, \method's design, which captures temporal context over different ranges, specializes the method for TAL and untrimmed videos, leading to its significant outperformance.
\begin{table*}[!ht]
\centering
\setlength{\tabcolsep}{4pt}
\resizebox{0.88\textwidth}{!}{%
\begin{tabular}{c} 
\begin{tabular}{l|c|
    c|c|c|ccccc|c}
    \hline
    
    \hline
       \multirow{2}{*}{Method}  & \multicolumn{2}{c|}{Backbone} & 
        \multirow{2}{*}{Flow}  & \multirow{2}{*}{\shortstack[c]{GPU\\(GB)}}  & \multicolumn{5}{c|}{mAP} & \multirow{2}{*}{Avg. mAP ($\uparrow$)}  \\ \cline{2-3} \cline{6-10}
       & Type & E2E Adaptation & & & 0.3 & 0.4 & 0.5 & 0.6 & 0.7 & \\\hline
       AFSD-RGB~\cite{lin2021learning} & I3D & 
       \xmark & \xmark & - & 57.7 & 52.8 & 45.4 & 34.9 & 22.0 & 43.6\\
       G-TAD~\cite{xu2020g} & TSN & 
       \xmark  & \cmark  & - & 54.5 & 47.6 & 40.3 & 30.8 & 23.4 & 39.3 \\
       TadTR~\cite{liu2022end} & I3D  & 
       \xmark & \cmark & - & 62.4 & 57.4 & 49.2 & 37.8 & 26.3 & 46.6\\
       TadTR~\cite{liu2022end} & SlowFast-101 & 
 
       \xmark  & \xmark & - & 70.4 & 66.4 & 58.3 & 46.8  & 33.5 & 55.1 \\
       AFSD~\cite{lin2021learning} & I3D  & 
       \xmark & \cmark & - & 67.3 & 62.4 & 55.5 & 43.7 & 31.1 & 52.0\\
       ActionFormer~\cite{zhang2022actionformer} & I3D  & 
       \xmark  & \cmark & -  & 82.1 & 77.8 & 71.0 & 59.4 & 43.9 & 66.8 \\
       Tridet~\cite{shi2023tridet} & I3D  & 
       \xmark  & \cmark & -  &  83.6 & 80.1 & 72.9 & 62.4 & 47.4 & 69.3  \\\hline
       E2E-TAD~\cite{liu2022empirical} & SlowFast-101 & 
       \cmark & \xmark & 14 & 71.4   & 66.6 & 59.4 & 48.1 & 36.8 & 56.4  \\
       TALLFormer~\cite{cheng2022tallformer} & VSwin-Base~\cite{liu2022video} & 
       \cmark & \xmark & 29 & 76.0 & - & 63.2 & - & 34.5 & - \\ 
       Re$^2$TAL~\cite{zhao2023re2tal} & Re$^2$VideoSwin-T & 
       \cmark & \xmark & 6.8 & 77.0 & 71.5 & 62.4 & 49.7 & 36.3 & 59.4 \\
       Re$^2$TAL~\cite{zhao2023re2tal} & Re$^2$SlowFast-101 & 
       \cmark & \xmark & 6.8 & 77.4 & 72.6 & 64.9 & 53.7 & 39.0 & 61.5 \\
       \hline \hline
       TadTR~\cite{liu2022end} & \multirow{2}{*}{SlowFast-101} & 
       \xmark  & \xmark & - & 70.4 & 66.4 & 58.3 & 46.8  & 33.5 & 55.1 \\
       \textbf{\method~(Ours)} &  &
       \cmark  & \xmark &  3.5 & 74.2 & 69.3 & 61.2 & 49.6 & 36.3 & 58.2{\tiny{\textcolor{teal}{$\mathord{\uparrow}3.1$}}} \\ \hline
       ActionFormer~\cite{wang2023videomae} & \multirow{2}{*}{\shortstack[c]{VideoMAEv2\\(ViT-Base)}} & 
       \xmark  & \xmark & - & 80.8 & 75.6 & 68.3 & 59.0 & 45.6 & 65.9 \\
       \textbf{\method~(Ours)} &  &
       \cmark  & \xmark & 6.5  & 81.1 & 77.0 & 70.2 & 61.1 & 46.9 & 67.3{\tiny{\textcolor{teal}{$\mathord{\uparrow}1.4$}}} \\ \hline
        ActionFormer~\cite{wang2023videomae} & \multirow{2}{*}{\shortstack[c]{VideoMAEv2\\(ViT-g)}} & 
        \xmark  & \xmark & - & 84.0  & 79.6 & 73.0 & 63.5 & 47.7 & 69.6 \\
        \cellcolor[gray]{.95}\textbf{\method~(Ours)} &   & 
        \cellcolor[gray]{.95}\cmark  & \cellcolor[gray]{.95}\xmark &   \cellcolor[gray]{.95}40.6 & \cellcolor[gray]{.95}\textbf{85.0} & \cellcolor[gray]{.95}\textbf{81.1} & \cellcolor[gray]{.95}\textbf{74.5} & \cellcolor[gray]{.95}\textbf{65.1} & \cellcolor[gray]{.95}\textbf{49.3} & \cellcolor[gray]{.95}\textbf{71.0}{\tiny{\textcolor{teal}{$\mathord{\uparrow}1.4$}}} \\
        \hline 

        \hline
    \end{tabular} \\
(a) \\
\begin{tabular}{l|c|
    c|c|c|ccc|c}
    \hline
    
    \hline
       \multirow{2}{*}{Method}  & \multicolumn{2}{c|}{Backbone} & 
       \multirow{2}{*}{Flow}  & \multirow{2}{*}{\shortstack[c]{GPU\\(GB)}}  & \multicolumn{3}{c|}{mAP} & \multirow{2}{*}{Avg. mAP ($\uparrow$)}  \\ \cline{2-3} \cline{6-8}
       & Type & E2E Adaptation & & & 0.5 & 0.75 & 0.95 & \\ \hline
       AFSD-RGB~\cite{lin2021learning} & I3D & 
       \xmark & \xmark & - & - & - & - & 32.9 \\
       G-TAD~\cite{xu2020g} & TSN & 
       \xmark  & \cmark  & - & 50.4 & 34.6 & 9.0 & 34.1 \\
       TadTR~\cite{liu2022end} & I3D  & 
       \xmark & \cmark & - & 49.1 & 32.6 & 8.5 & 32.3\\
       AFSD~\cite{lin2021learning} & I3D  & 
       \xmark & \cmark & - & 52.4 & 35.3 & 6.5 & 34.4\\
       ActionFormer~\cite{zhang2022actionformer} & I3D  & 
       \xmark  & \cmark & -  & 53.5 & 36.2 & 8.2 & 35.6 \\ 
       Tridet~\cite{shi2023tridet} & R(2+1)D  & 
       \xmark  & \cmark &  - &  54.7 & 38.0 & 8.4 & 36.8  \\\hline
       E2E-TAD~\cite{liu2022empirical} & SlowFast-50  & 
       \cmark  & \xmark & 14  & 50.5 & 36.0 & 10.8 & 35.1 \\ 
       TALLFormer~\cite{cheng2022tallformer} & VSwin-Base~\cite{liu2022video} & 
       \cmark & \xmark & 29 & 54.1 & 36.2 & 7.9 & 35.6 \\ 
       Re$^2$TAL~\cite{zhao2023re2tal} & Re$^2$VideoSwin-T & 
       \cmark & \xmark & 6.8 & 54.75 & 37.81 & 9.03 & 36.8 \\
       Re$^2$TAL~\cite{zhao2023re2tal} & Re$^2$SlowFast-101 & 
       \cmark & \xmark & 6.8 & 55.3 & 37.9 & 9.1 & 37.0 \\
       \hline \hline
       ActionFormer~\cite{wang2023videomae} & \multirow{2}{*}{\shortstack[c]{VideoMAEv2\\(ViT-Base)}} & 
       \xmark  & \xmark & - & 56.5 & 37.8 & 7.7 & 36.8 \\
       \textbf{\method~(Ours)} &  &
       \cmark  & \xmark &  6.5 & 57.7 & 38.6 & 8.1 &  38.1{\tiny{\textcolor{teal}{$\mathord{\uparrow}1.3$}}} \\ \hline
        ActionFormer~\cite{wang2023videomae} & \multirow{2}{*}{\shortstack[c]{VideoMAEv2\\(ViT-g)}} & 
        \xmark  & \xmark & - & 57.2 & 38.3 & 5.8 & 37.1 \\
        \cellcolor[gray]{.95}\textbf{\method~(Ours)} &   &
        \cellcolor[gray]{.95}\cmark  & \cellcolor[gray]{.95}\xmark & \cellcolor[gray]{.95}40.6 & \cellcolor[gray]{.95} 58.5 & \cellcolor[gray]{.95} 39.8 & \cellcolor[gray]{.95} 7.8 & \cellcolor[gray]{.95}\textbf{38.6}{\tiny{\textcolor{teal}{$\mathord{\uparrow}1.5$}}} \\
        \hline 

        \hline
    \end{tabular} \\
(b) \\
\end{tabular}
}
\vspace{-3mm}
\caption{Temporal action localization performance comparison of \method with state-of-the-art methods on (a) \textbf{THUMOS-14} and (b) \textbf{ActivityNet-v1.3}. E2E Adaptation is \cmark~when backbone adaptation~(end-to-end training) happens along with learning the TAL head. Flow is \cmark~when optical flow features are used. GPU represents peak GPU memory in GBs occupied for training for batch size of 1 on an A100 GPU. GPU is `-' when the backbone is frozen~(\ie E2E Adaptation is \xmark). \method can significantly outperform all existing methods including those using both RGB and Flow as well as performing backbone adaptation for TAL. 
}
\label{tab:main_thumos_anet}
\vspace{-5mm}
\end{table*}
\subsection{Comparison with State-of-the-Art}
We compare \method on THUMOS-14 and ActivityNet-v1.3 with existing TAL methods in Tables~\ref{tab:main_thumos_anet}a and \ref{tab:main_thumos_anet}b. Given the diversity of different setups in previous approaches, we include columns mentioning the video backbone used, whether the method uses optical flow features, and whether the training involves head-only transfer learning~(\xmark~in~E2E Adaptation column) or backbone adaptation/end-to-end training~(\cmark~in E2E Adaptation column). We also include a column to provide the peak GPU memory utilization of adapting the backbone during training with a batch size of 1 on an A100 GPU~(for head-only with no backbone adaptation, we report it as -). 

As per Table~\ref{tab:main_thumos_anet}a, we can observe that \method significantly outperforms head-only training~(\xmark~in~E2E Adaptation column) on both CNN-based and transformer-based video backbones -- SlowFast-101, VideoMAEv2 (ViT-B), and VideoMAEv2 (ViT-g), by $3.1\%$, $1.4\%$, and $1.4\%$ on Avg mAP respectively. Similarly in Table~\ref{tab:main_thumos_anet}b, we can see that \method outperforms head-only transfer learning on all the video backbones -- VideoMAEv2 (ViT-B) and VideoMAEv2 (ViT-g), by $1.3\%$ and $1.5\%$ on Avg mAP respectively. This shows the effectiveness of \method in better leveraging and adapting video backbones across different sizes for TAL beyond head-only transfer learning. \method outperforms all existing TAL methods, including those that use both RGB and optical flow features and those that attempt backbone adaptation, thereby establishing a new SOTA on both THUMOS-14 and ActivityNet-v1.3.

\begin{table*}[h]
\centering
\resizebox{\textwidth}{!}{%
\begin{tabular}{ccc} 
\begin{tabular}{lc}
\hline
\multirow{2}{*}{\textbf{Setup}} & \multirow{2}{*}{\shortstack[c]{\bf Avg\\\bf mAP}}
\\
& \\
 \hline
\method without Long-range Adapter & 70.2 \\
\method without Short-range Adapter & 70.3 \\
\method without Long-Short-range Gated Fusion & 69.8  \\
\method (Ours) & \textbf{71.0}\\ \hline 

\hline
\end{tabular}
&
\begin{tabular}{lc}
\hline
\multirow{2}{*}{\textbf{Gating Strategy}} & \multirow{2}{*}{\shortstack[c]{\bf Avg\\\bf mAP}}
\\
& \\
 \hline
Gating with random initialization  & 70.0 \\
Gating with one initialization & 69.8 \\
Gating with zero initialization (Ours) & \textbf{71.0} \\ 
 \hline 

\hline
\end{tabular}
&
\begin{tabular}{lc}
\hline
\multirow{2}{*}{\shortstack[c]{\bf Intermediate Layers\\\bf($f_i$)}} & \multirow{2}{*}{\shortstack[c]{\bf Avg\\\bf mAP}}
\\
& \\
 \hline 
$f_1, \hdots, f_{20}$ &  44.4 \\
$f_{21}, \hdots, f_{39}$ & 70.6 \\
$f_{10}, \hdots, f_{30}$ & 69.4  \\
$f_{30}, \hdots, f_{39}$ & 69.9 \\
$f_{15}, \hdots f_{20}$, $f_{35}, \hdots, f_{39}$ & 68.9 \\
$f_1, \hdots, f_{39}$~(All, Ours) & \textbf{71.0}\\ \hline 

\hline
\end{tabular} \\
(a) & (b) & (c)
\end{tabular}
}
\vspace{-4mm}
\caption{(a) Ablation showing the effectiveness of each component of \method. (b) Comparison with different gating strategies showing the significance of doing zero initialization for achieving the best Avg mAP. (c) Analysis on attaching spatial and temporal adapters to different sets of intermediate layers. All experiments in (a-c) performed on THUMOS-14 using VideoMAEv2~(ViT-g).}
\label{tab:ablation_gating_layers}
\vspace{-6mm}
\end{table*}
\subsection{Ablation}
\looseness=-1 We conduct an ablation study using THUMOS-14 and VideoMAEv2~(ViT-g), as shown in Table~\ref{tab:ablation_gating_layers}a, to highlight the effectiveness of each integral component of \method. The table shows that omitting the Long-range or Short-range Adapter reduces Avg mAP by at 
least $0.8\%$, indicating the necessity of both in incorporating temporal information at different ranges from the intermediate layers for optimal TAL performance. Next, we conduct an ablation where we remove the Long-Short-range Gated Fusion module and replace it with a simple addition of the features. As Row 3 in Table~\ref{tab:ablation_gating_layers}a shows, removing the Long-Short-range Gated Fusion module leads to a significant drop of 
$1.2\%$ in Avg mAP. This shows that along with adapting the long-range and short-range temporal information via the respective adapters, it is also critical to learn how to scale the contribution across the intermediate layers to allow the most relevant long-range and short-range temporal information to be incorporated into the features being fed to the TAL head.

\subsection{Discussion}
\looseness=-1 \noindent{\bf Gating with zero initialization outperforms all other gating strategies.}
In comparing \method's gating layer strategy with alternatives, our focus is on its unique design and efficiency. As detailed in Sec~\ref{sec:loshrf}, we enforce a zero initialization on the scaling parameter in the long-range and short-range gating layers to enable the long-range and short-range contributions to function as a residual contribution with respect to the last layer features.  Table~\ref{tab:ablation_gating_layers}b shows that this approach yields the highest Avg mAP, outperforming random or one-value initializations. 

\looseness=-1 \noindent{\bf Gating parameter learns different values over intermediate layers.} Since the scaling parameter in the long-range and short-range gating layers is a learnable parameter, we assess the distribution of the values learned by the scaling parameter post-training across the intermediate layers of the video backbone. For VideoMAEv2~(ViT-g) on THUMOS-14, we find that the value of learnable parameter ranges in [-0.01, 0.52] across the long-range and short-range gating in the intermediate layers. This validates two hypotheses. One, all learnable parameters do not collapse trivially to their originally initialized value of 0. At least some of them learn a non-zero value to scale and provide a meaningful residual contribution from intermediate layers to the TAL-enhanced features, $\mathbf{FT^{X}_N}$, entering the TAL head. Two, the learnable parameters exhibit a range of values which indicates that the scaling parameters learn to contribute differently from the intermediate layers as per their importance in improving $\mathbf{FT^{X}_N}$.

\noindent{\bf Effect of adapting different intermediate layers.} Our experiments with VideoMAEv2~(ViT-g) on THUMOS-14 reveal that attaching Long-range and Short-range Adapters to all 40 transformer layers yields the highest Avg mAP (see Table~\ref{tab:ablation_gating_layers}c), underscoring the contribution of each layer's temporal information over different time spans to the TAL head in enhancing localization performance. We can also observe that with the deeper half layers, $f_{21}, \hdots, f_{39}$, we get very close to optimal Avg mAP while shallower half layers, $f_{1}, \hdots, f_{20}$, leads to low Avg mAP. We believe this is due to deeper layers capturing more comprehensive information about the video than shallower layers. While a combination of all layers is optimal, using only the deeper half layers can suffice when training resources are limited.

\noindent{\bf Significant gains on ActivityNet-v1.3.} Tab.~\ref{tab:main_thumos_anet} shows that improvements between successive works in recent years on ActivitNet-v1.3 is generally around $0.5-1\%$ Avg mAP.
In contrast, \method can improve by more than 1\% Avg mAP compared to any existing baseline, showing the significant performance gain achieved by \method on ActivityNet-v1.3. 

\vspace{-4mm}
\begin{figure}[h]
    \centering
    \includegraphics[width=0.75\columnwidth]{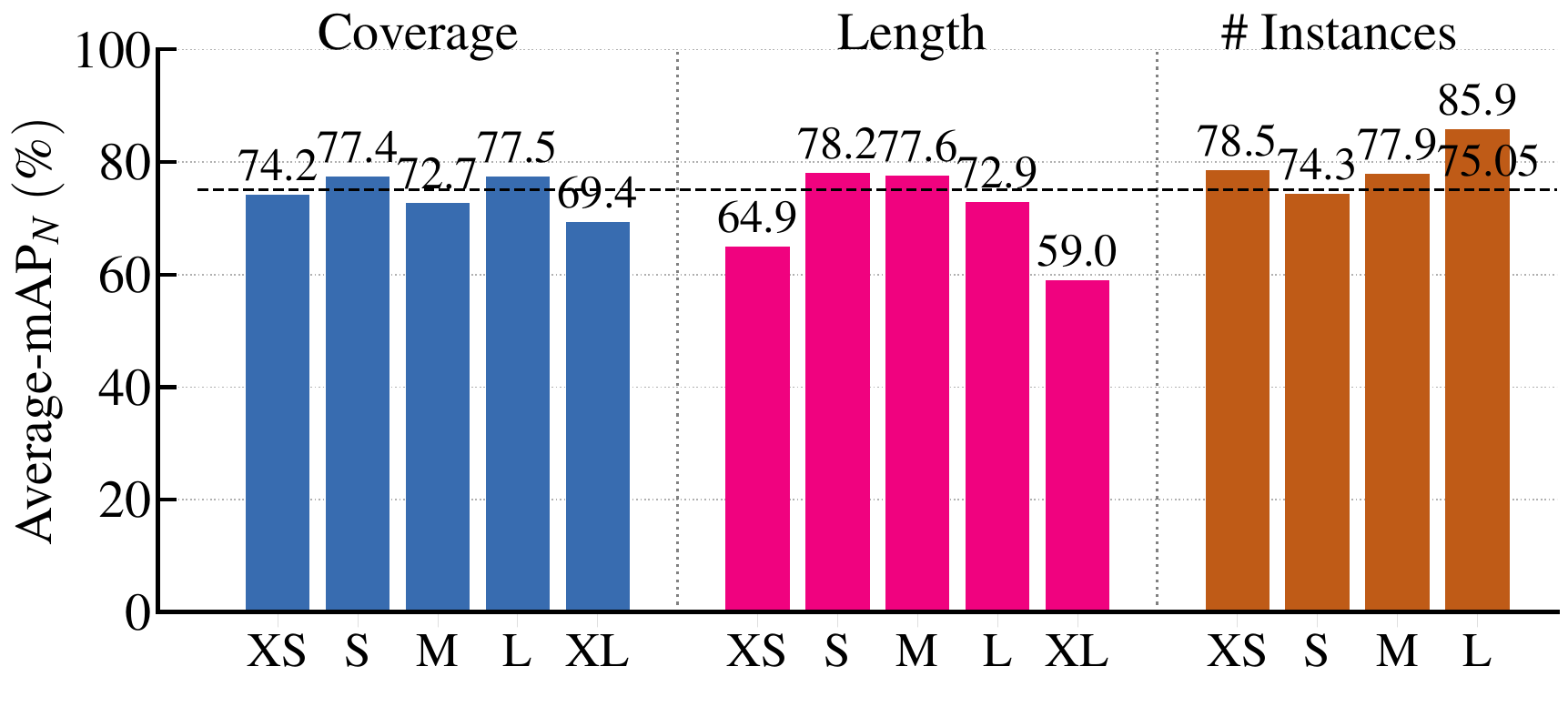}
    \vspace{-3mm}
    \includegraphics[width=0.75\columnwidth]{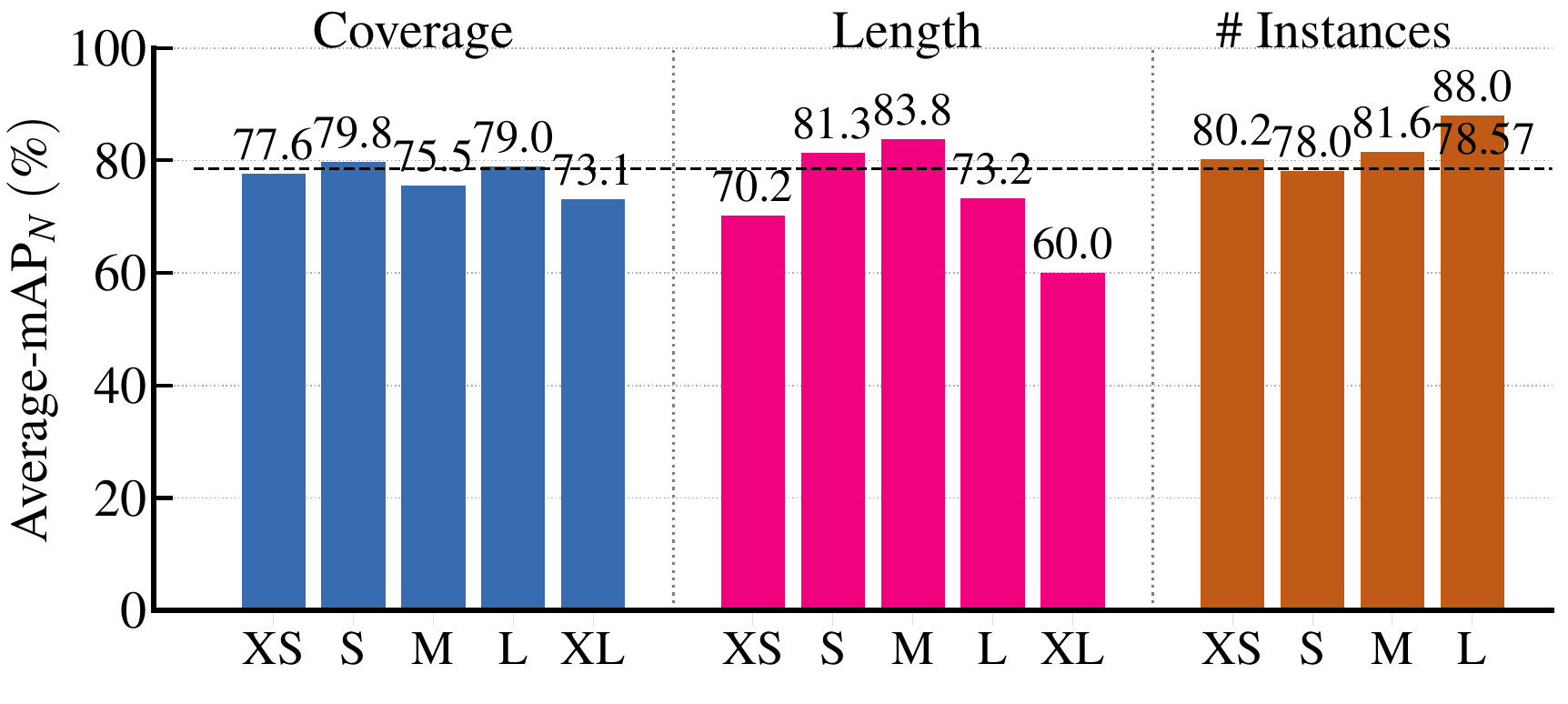}
    \vspace{-2mm}
    \caption{Sensitivity analysis on THUMOS-14 using \cite{alwassel2018diagnosing}. mAP$_N$ denotes normalized mAP at tIoU=0.5 with N average ground truth segments per class. Top: \method w/o Long-Short-Adapter. Bottom: \method~(Ours).  Performance for both XS and XL improves significantly with our method \method~(bottom) compared to the baseline, \method w/o Long-Short-range Adapter~(top).}
    \vspace{-5mm}
    \label{fig:visual}
\end{figure}
\noindent{\bf LoSA improves both long- and short-action instances.} Fig~\ref{fig:visual} shows a sensitivity analysis on THUMOS-14 using VideoMAEv2~(VIT-g), taking into account coverage, length, and number of action instances. We show a comparison between \method w/o Long-Short-range Adapter~(baseline, \textit{the top row}) and \method~(ours, \textit{the bottom row}). Refer to supplementary for setup details.
We observe that the performance for both XS and XL improves significantly with our method \method compared to the baseline, \method w/o Long-Short-range Adapter. We believe this is because \method enables untrimmed temporal learning at different temporal ranges in the intermediate layers via the Long- and Short-range Temporal Adapters. This enables capturing complex scene details as well as fine-grained information required for long- and short- duration action instances respectively.
\vspace{-2mm}
\section{Conclusion}
We introduce \method, the first memory-and-parameter-efficient backbone adapter designed specifically for TAL. \method comprises a novel design of Long-range and Short-range Temporal Adapters that are attached to the intermediate layers to adapt them towards improving TAL, and run parallel to the video backbone to reduce memory footprint. Finally, Long-Short-range Gated Fusion module takes the output from these adapters to fuse and give TAL-enhanced features. 
This allows \method to scale end-to-end backbone adaptation to $>$1\,B parameter backbones like VideoMAEv2 (ViT-g) and leverage them beyond head-only training to significantly outperform all existing TAL methods. Our work is the first to go beyond traditional techniques, including full model adaptation and head-only transfer learning, in addressing the challenging problem of adapting video backbones for end-to-end TAL, and proves effective in leveraging large foundation models for 
TAL in untrimmed videos. 
\section{Acknowledgment}

Resources used in preparing this research were provided by Microsoft, and, in part, by the Province of Ontario, the Government of Canada through CIFAR, and  \href{http://www.vectorinstitute.ai/partnerships/current-partners/}{partners of the Vector Institute}.

{\small
\bibliographystyle{ieee_fullname}
\bibliography{egbib}
}

\renewcommand{\thesection}{S\arabic{section}} 

\setcounter{table}{0}
\renewcommand{\thetable}{S\arabic{table}}

\setcounter{figure}{0}
\renewcommand{\thefigure}{S\arabic{figure}}

\setcounter{section}{0}
\section{Additional Results}

In this document, we provide additional analysis and details for our work \method. Section~\ref{sec:supp_visuals} provides qualitative analysis of \method by visualizing and comparing the action snippets localized in the videos. 
Section~\ref{sec:supp_analysis} provides error analysis of \method to highlight additional aspects of the method. 
Finally, Section~\ref{sec:supp_limit} expands on the limitations and future work of the \method.

\section{Visualizations}
\label{sec:supp_visuals}
\begin{figure}
\centering
 \includegraphics[width=0.99\textwidth,height=0.99\textheight,keepaspectratio]{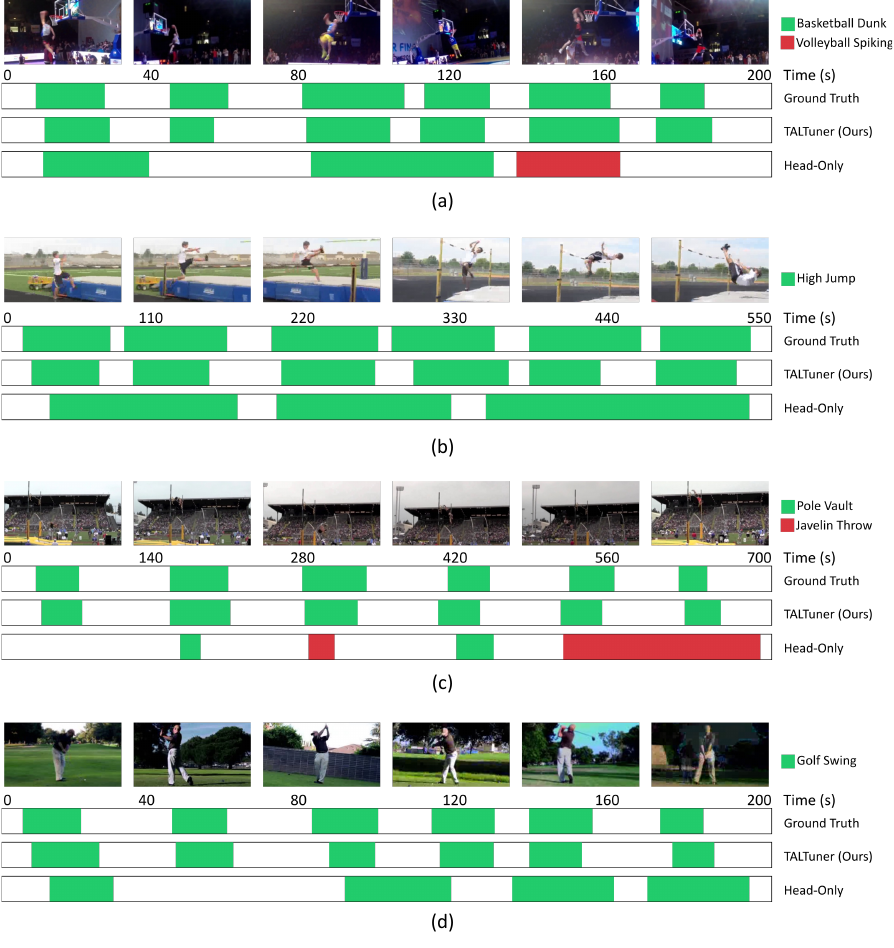}
 \caption{Visualizations of \method vs.~baseline~(Head-only Transfer Learning) for THUMOS-14 on VideoMAEv2~(ViT-g). Across all the visualizations~(a-d), \method is able to localize action snippets~(in \textcolor{green}{green}) with action boundaries significantly closer to the ground truth than the baseline, leading to fewer false positives and false negatives. \method also predicts the action class for the snippets more accurately than the baseline~(seen by incorrect class predictions in \textcolor{red}{red} by the baseline in (a) and (c)).}
 \label{fig:vis_thumos}
\end{figure}

In Fig~\ref{fig:vis_thumos}, we provide additional visualizations of the action snippets localized by \method compared to the baseline of head-only transfer learning in videos from THUMOS-14 using VideoMAEv2~(ViT-g). We can observe that across all the visualizations~(Fig~\ref{fig:vis_thumos}a-d), \method is able to localize action snippets with action boundaries significantly closer to the ground truth than the baseline while also predicting the action class for the snippets more accurately than the baseline. Fig~\ref{fig:vis_thumos}a shows a video of ``Basketball Dunk''. We can observe that, compared to head-only, \method is able to localize the action boundaries for ``Basketball Dunk'' more precisely with respect to the ground truth. 
We believe this is due to \method's ability to induce untrimmed temporal video understanding at different temporal ranges in the intermediate layers via the long-range and short-range adapters. This enhances the informativeness of the adapted features of the intermediate layers, contributing towards directly improving TAL and allows to make fine distinctions between foreground and background around action boundaries. 
This effect is further visible around 160\,s, where \method correctly predicts the snippet action but head-only, due to insufficient temporal context, misclassifies the action as ``Volleyball Spiking'', which has similar temporal motion as ``Basketball Dunk''.  

\begin{figure}
 \includegraphics[width=\textwidth,height=\textheight,keepaspectratio]{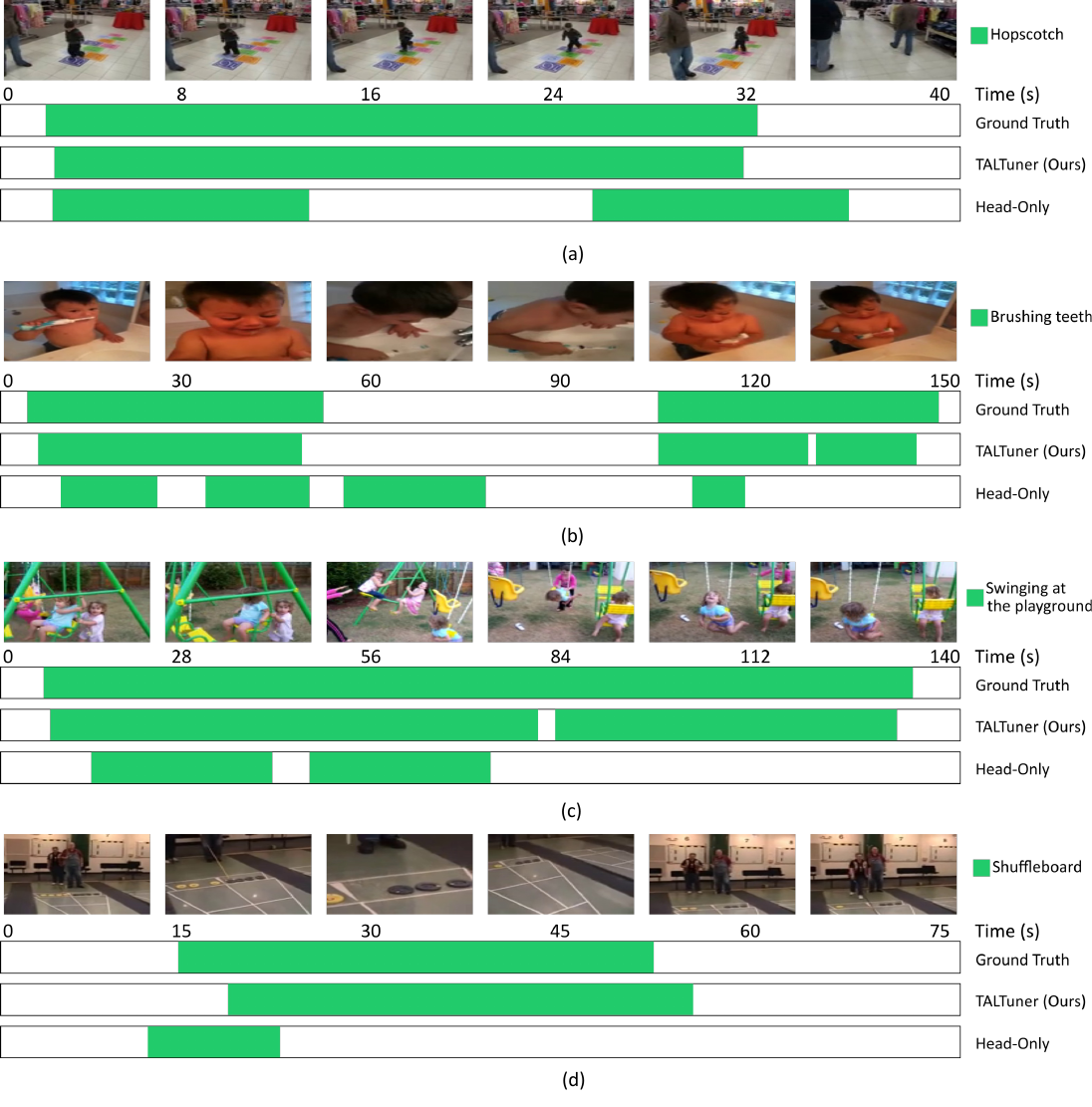}

 \caption{Visualizations of \method vs.~baseline~(Head-only Transfer Learning) for ActivityNet-v1.3 on VideoMAEv2~(ViT-g). Across all the visualizations~(a-d), \method is able to localize action snippets~(in \textcolor{green}{green}) with action boundaries significantly closer to the ground truth than the baseline, leading to fewer false positives and false negatives.}
 \label{fig:vis_anet}
\end{figure}

In Fig~\ref{fig:vis_anet}, we provide visualizations of the action snippets localized by \method compared to the baseline of head-only transfer learning in videos from ActivityNet-v1.3 using VideoMAEv2~(ViT-g). We can observe that across all the visualizations~(Fig~\ref{fig:vis_anet}a-d), \method is able to localize action snippets with action boundaries significantly closer to the ground truth than the baseline. In Fig~\ref{fig:vis_anet}a, where the video shows a kid playing Hopscotch, while the baseline misses the action between 16-24s~(false negative) and incorrectly predicts the background as action between 32-40s~(false positive), \method is able to mitigate both false negative and false positive and accurately predict the start and end timestamps of the action. 
We believe that this is due to \method's ability to induce untrimmed temporal video understanding at different temporal ranges in the intermediate layers via the long-range and short-range adapters. This improves the adapted feature sequence at each intermediate layer with respect to TAL, allowing the TAL head to perform better action localization.
\label{sec:visualization}

\clearpage
 \onecolumn

\clearpage

\clearpage

\twocolumn

\section{Additional Analysis}
\label{sec:supp_analysis}

\begin{figure}[t]
    \centering
    \includegraphics[width=\columnwidth]{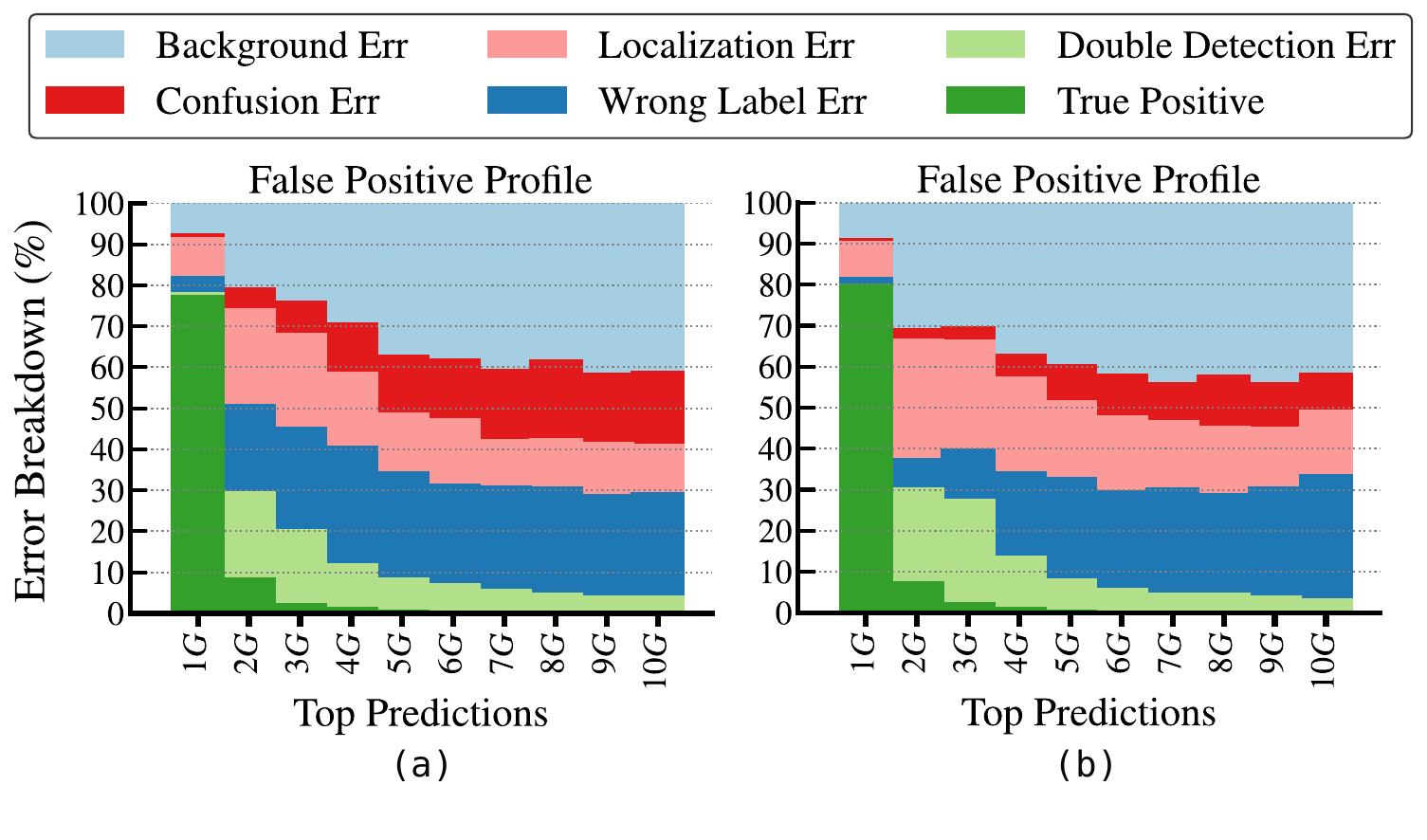}
    \caption{False positive~(FP) profiling on THUMOS-14 using~\cite{alwassel2018diagnosing}. FP error breakdown for top-10 ground-truth~(GT) predictions comparing (a) \method w/o Long-Short-range Adapter and (b) \method~(ours). Wrong label prediction error significantly drops with \method compared to \method w/o Long-Short-range Adapter. }
    \label{fig:false_positive_thumos}
\end{figure}

In Fig~\ref{fig:false_positive_thumos}, we conduct a False Positive (FP) analysis at tIoU=0.5 for THUMOS-14 using VideoMAEv2~(VIT-g). 
We show comparison between the baseline, \method w/o Long-Short-range Adapter~(Fig~\ref{fig:false_positive_thumos}a) and our method \method~(Fig~\ref{fig:false_positive_thumos}b). We can see a drop in the wrong label prediction error with \method compared to \method w/o Long-Short-range Adapter. This shows the significance of incorporating untrimmed temporal video understanding while adapting the intermediate layers for TAL. The chart shows FP error breakdown for top-10 ground truth~(GT) predictions. For more details regarding the chart, we refer the readers to \cite{alwassel2018diagnosing}.

\section{Limitations, Negative Impact, and Future Work}
\label{sec:supp_limit}
To our best knowledge, we do not perceive a potential negative impact that is specific to our proposed method. While \method's memory-efficient design allows to leverage billion-parameter-plus models like VideoMAEv2~(ViT-g) for end-to-end TAL, the memory requirement is still linearly dependent~(asymptotically) on the number of frames, frame resolution, and model size to a certain degree. In future, we can explore reducing the memory usage to sub-linear while continuing to improve performance as we leverage larger foundation models. Further interesting directions include extending to end-to-end spatio-temporal localization, end-to-end video object segmentation, end-to-end video grounding, and other multi-modal video understanding tasks involving audio, text, and other modalities.

\end{document}